\pdfoutput=1

\documentclass[11pt]{article}

\usepackage[preprint]{template/acl}

\usepackage{times}
\usepackage{latexsym}

\usepackage[T1]{fontenc}

\usepackage[utf8]{inputenc}

\usepackage{microtype}

\usepackage{inconsolata}

\usepackage{graphicx}

\usepackage{float}
\usepackage{booktabs}
\usepackage{cleveref}
\usepackage{multirow}
\usepackage{hyperref}

\title{Command R7B Arabic: A Small, Enterprise Focused, Multilingual, and Culturally Aware Arabic LLM}

\author{
 \textbf{Yazeed Alnumay\thanks{Equal contribution. Authors appear in alphabetical order by second name.}},
 \textbf{Alexandre Barbet\textsuperscript{*}},
 \textbf{Anna Bialas\textsuperscript{*}},
 \textbf{William Darling\textsuperscript{*}},
\\
 \textbf{Shaan Desai\textsuperscript{*}},
 \textbf{Joan Devassy\textsuperscript{*}},
 \textbf{Kyle Duffy\textsuperscript{*}},
 \textbf{Stephanie Howe\textsuperscript{*}},
\\
 \textbf{Olivia Lasche\textsuperscript{*}},
 \textbf{Justin Lee\textsuperscript{*}},
  \textbf{Anirudh Shrinivason\textsuperscript{*}},
   \textbf{Jennifer Tracey\textsuperscript{*}}
\\
\\
 Cohere
}

\begin{document}

\maketitle

\begin{abstract}
Building high-quality large language models (LLMs) for enterprise Arabic applications remains challenging due to the limited availability of digitized Arabic data. In this work, we present a data synthesis and refinement strategy to help address this problem, namely, by leveraging synthetic data generation and human-in-the-loop annotation to expand our Arabic training corpus. We further present our iterative post training recipe that is essential to achieving state-of-the-art performance in aligning the model with human preferences, a critical aspect to enterprise use cases. The culmination of this effort is the release of a small, 7B, open-weight model that outperforms similarly sized peers in head-to-head comparisons and on Arabic-focused benchmarks covering cultural knowledge, instruction following, RAG, and contextual faithfulness.
\end{abstract}

\section{Introduction}
Multilingual language models are evolving rapidly \citep{multilingual_llm_survey}, yet specific languages and capabilities remain underdeveloped, particularly in enterprise applications. While state-of-the-art models continue to improve, they often struggle to adapt to linguistic and professional needs in languages like Arabic \cite{cdt2023}. This challenge becomes even more pronounced when additional constraints are introduced: the need to keep the model small to ensure accessibility even with limited resources, overcoming data scarcity, and accounting for linguistic nuances that do not translate well from English, all the while prioritizing rapid iteration to stay aligned with the fast-moving market. To address these issues, we developed a post-training approach that efficiently tailors cutting-edge models to specialized capabilities. This report outlines our methodology and findings, offering insights into adapting LLMs for language-specific and professional domains.

\section{Related Work}
With the recent rapid development in LLMs \citep{llm_survey}, some focus was placed on improving model multilingualism through second language acquisition techniques \citep{multilingual_llm_survey}. These techniques aim to circumvent data scarcity in languages other than English by adding other language capabilities to English models, which is more data efficient. For instance, the Llama 3 family of models adds a final pretraining stage by adding multilingual pretraining data mixed with English \citep{llama3}. These techniques have been applied to Arabic-centric models, such as ALLaM \citep{allam}, Jais \citep{jais, jais_family}, AceGPT \citep{acegpt1, acegpt2, acegpt3}, and Fanar \citep{fanar}. These projects primarily focused on pretraining data mixture, staging, and tokenizer innovations, including vocabulary expansion (ALLaM), iterative vocabulary expansion (AceGPT), and morphology-based tokenization (Fanar). While they contribute strong foundational models for the community, they do not offer computationally efficient post-training methods.

Post-training has become essential for building robust models \cite{sft,kumar2025llmposttrainingdeepdive,ouyang2022traininglanguagemodelsfollow}. Many research labs have contributed to the open-source community by documenting modern post-training techniques. Notable examples include Tülu 3 \cite{lambert2025tulu3pushingfrontiers}, which provides a comprehensive overview of general post-training methods, and Aya Expanse \cite{aya_expanse}, which focuses on multilingual adaptation.

\begin{figure*}[t]
    \centering
    \includegraphics[width=0.9\linewidth]{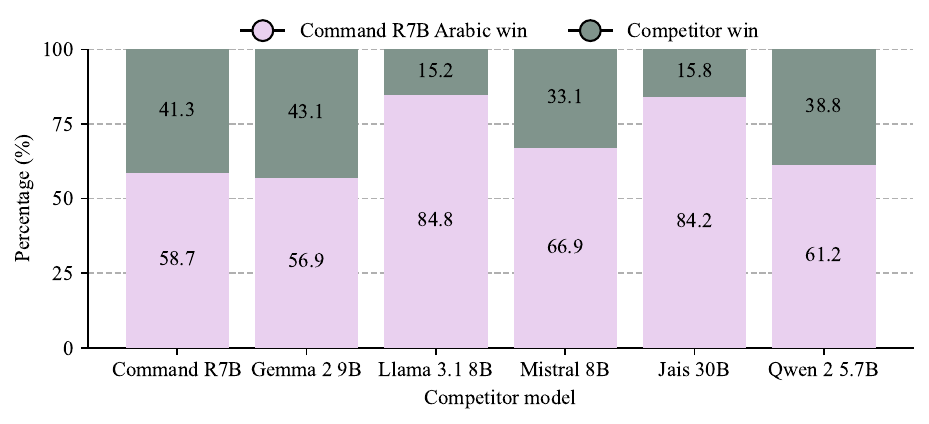}
    \caption{Evaluations on enterprise usability factors (mArenaHard, described in \Cref{sec:results}). Auto win-rates on Arabic version of LMSYS Arena "Hard" human preference tasks \citep{aya_expanse}. Command R7B Arabic outperforms all listed similarly-sized models.}
    \label{fig:model_head_to_head}
\end{figure*}

Our work builds on these efforts by developing a systematic, iterative, and comprehensive approach to efficiently adapt LLMs for languages. Specifically, we leverage iterative tuning \cite{llama3} methods that rely on best-of-N sampling to generate instruction and preference data via automated reward models or human preference \citep{yuan2024selfrewardinglanguagemodels}. We also further reduce compute requirements by incorporating model merging techniques \citep{mergekit, model_merging}.

\section{Methods}
\begin{figure*}[tb]
    \centering
    \includegraphics[width=0.9\linewidth]{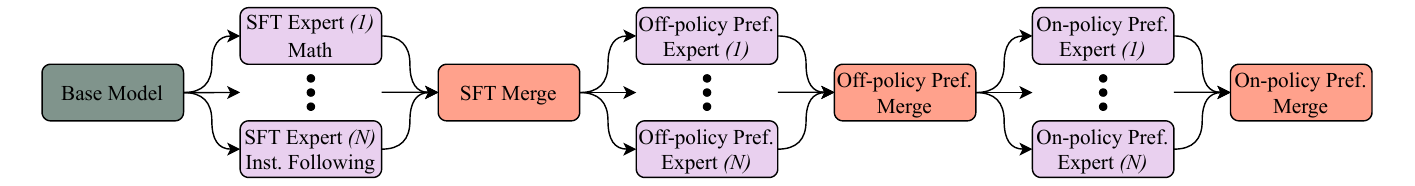}
    \caption{Outline of Command R7B Arabic's training processes with three training stages, each training multiple experts that are merged into a single general model. For instance, in the SFT stage, multiple SFT expert models are trained to excel in specific domains, such as mathematics or instruction following. These experts are subsequently merged to create a generalist SFT model via parameter-wise linear interpolation of the experts' weights.}
    \label{fig:training_steps}
\end{figure*}

Our training procedure is illustrated in \Cref{fig:training_steps}. We start by selecting a strong starting model (\Cref{subsec:base_model_selection}), on which we perform three distinct training phases: \emph{(i)} supervised fine-tuning (SFT) \citep{sft}, for which we employ iterative dataset refinement techniques (\Cref{subsec:multilingual_arbitrage,subsec:iterative_supervised_refinement}), \emph{(ii)} off-policy (offline) preference tuning, and \emph{(iii)} iterative preference tuning. The latter two are described in \Cref{subsec:preference_tuning}. After each training phase, we merge expert models into a single general model (\Cref{subsec:model_merging}).

\subsection{Base Model Selection}
\label{subsec:base_model_selection}
As a starting checkpoint, we chose Command R7B \citep{cr7bblog} - a strong, general purpose open-weight model already trained on a large corpus of multilingual data, including Arabic. Our primary objective was to reach state-of-the-art performance in Arabic enterprise use cases while preserving the model’s performance on other core capabilities. Starting from an already polished checkpoint meant we could spend more effort on our data and training efforts that refined Arabic-specific tasks. 

\subsection{Multilingual Arbitrage for Capability Enhancement}
\label{subsec:multilingual_arbitrage}
Previous work by Aya \citep{odumakinde2024multilingualarbitrageoptimizingdata} has demonstrated that synthetic data generation is crucial for achieving state-of-the-art performance, and this is especially true for domains with limited data availability such as Arabic. However, a key challenge when training Arabic LLMs is the distinctive difference between Arabic and English. Not only do these languages differ in syntax and morphology, but there are also variations in cultural and contextual nuances that make literal translation challenging. For example, lexical control tasks such as length adherence and structured generation are awkward or nonsensical when translated to Arabic.

To address this, we implemented a human-in-the-loop approach:

\begin{itemize}
    \item We collaborated with expert annotators to translate IFEval \citep{zhou2023instructionfollowingevaluationlargelanguage} instructions into Arabic. Additionally, we augmented the set with two instructions specific to the Arabic language: ``add $N$ diacritics to the response'' and ``use a specific grammatical verb to start sentences''. This ensured better alignment with Arabic linguistic and cultural nuances.
    \item These instructions were used as seeds to synthetically generate instruction following prompts in Arabic and subsequently the corresponding completions.
    \item In accordance with the work done in Aya’s Multilingual Arbitrage \cite{odumakinde2024multilingualarbitrageoptimizingdata}, we scored and filtered completions using a reward model, a panel of LLM judges for Arabic natural language quality, and max reward difference for preference pair dataset creation. 
\end{itemize}

This targeted approach ensured that the model learned to follow instructions naturally in Arabic, which is apparent in arena style win-rates where our model is consistently favored over other competitor models, as shown in \Cref{fig:model_head_to_head}.

\subsection{Dataset Curation and Iterative Supervised Refinement}
\label{subsec:iterative_supervised_refinement}

\begin{figure}[ht]
    \centering
    \includegraphics[width=\linewidth]{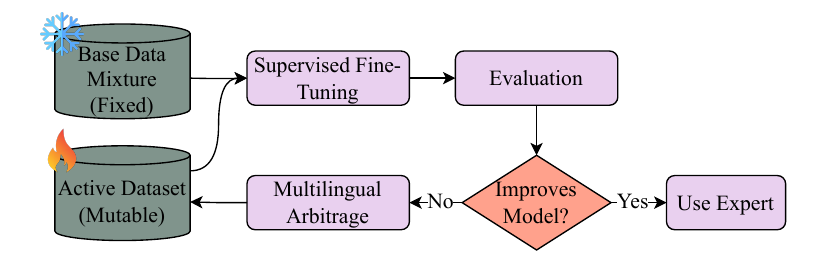}
    \caption{Flowchart for our iterative supervised refinement approach. It ensures that all datasets used improve targeted model performance by mixing a base data mixture with a targeted dataset that is iteratively improved via multilingual arbitrage.}
    \label{fig:iterative_supervised_refinement}
\end{figure}

The availability of high-quality Arabic datasets is a well-documented challenge \cite{cdt2023}. We aimed to incorporate both publicly available datasets, including ArMATH \cite{alghamdi-etal-2022-armath}, ArabicaQA \cite{abdallah2024arabicaqacomprehensivedatasetarabic}, and synthetically generated datasets, while enforcing a high-quality data standard. With this in mind, we defined the Iterative Supervised Refinement during Supervised Fine-Tuning (SFT) training phase as a process to optimize our dataset composition. The steps are illustrated in \Cref{fig:iterative_supervised_refinement} and are as follows:

\begin{enumerate}
    \item Define a base data mix consisting of high-quality instruction-tuning data.
    \item For each new dataset in consideration, add it to the base data mixture and fine-tune the model.
    \item Evaluate the resulting model using a benchmark evaluation harness to measure the impact of the new dataset.
    \item If the dataset improves performance in any critical capability, retain it for the next iteration.
    \item If no improvement was observed, apply Multilingual Arbitrage, refining the prompts before re-running the process.
\end{enumerate}

This approach enabled us to design an optimal dataset mixture that maximized the model’s instruction-following capabilities while maintaining a high standard for data quality. 

\subsection{Preference Tuning for Final Model Optimization}
\label{subsec:preference_tuning}
Since we initialized from a strong Command R7B model, it was essential to ensure that enhancements in Arabic did not degrade performance on other benchmarks. Similar to the methodology described by Aya \cite{ustun-etal-2024-aya}, we used two stages of preference tuning as final polishing steps to improve model performance and align it with human preferences. In the first phase, we performed offline preference training on general preference datasets to refine the model’s conversational fluency. In the second phase, we ran iterative preference training, incorporating an Arabic-translated reasoning and math-focused dataset \cite{alghamdi-etal-2022-armath}, which proved particularly beneficial for maintaining high performance across diverse enterprise use cases. Both preference tuning stages utilize the direct preference optimization (DPO) \citep{dpo} algorithm.

\begin{table*}[t]
  \centering
  \resizebox{\textwidth}{!}{%
    \begin{tabular}{lcccccc}
      \toprule
      \multirow{2}{*}{\textbf{Benchmark}} & \multirow{2}{*}{\textbf{R7B Arabic}} & \textbf{R7B} & \textbf{Gemma 9B} & \textbf{Llama 3.1 8B} & \textbf{Qwen 2.5 7B} & \textbf{Ministral 8B} \\
      & & \footnotesize\citep{cr7bblog} & \footnotesize\citep{gemma2} & \footnotesize\citep{llama3} & \footnotesize\citep{qwen2.5} & \footnotesize\citep{ministral} \\
    \midrule
      AlGhafa-Native & \textbf{82.2} & 81.5 & 81.3 & 80.1 & 80.2 & 76.6 \\
      ArabicMMLU & 60.9 & 59.7 & \textbf{62.4} & 56.6 & 61.2 & 53.6 \\
      IFEval AR & \textbf{69.0} & 57.8 & 67.8 & 48.4 & 62.4 & 49.3 \\
      TyDIQA-GoldP Arabic & \textbf{83.0} & 79.9 & 76.4 & 65.9 & 60.9 & 57.7 \\
      FaithEval Arabic & \textbf{51.6} & 49.9 & 47.0 & 40.9 & 49.9 & 25.5 \\
      \midrule
      \textbf{Average} & \textbf{69.3} & 65.8 & 67.0 & 58.4 & 62.9 & 52.5 \\
      \bottomrule \\
    \end{tabular}%
  }
  \caption{Full performance comparison against competitor models on Arabic-specific benchmarks. The highest score in each row is in \textbf{bold}. Command R7B Arabic is best-in-class compared to similarly sized models on all Arabic benchmarks, with the exception of ArabicMMLU.}
  \label{tab:arabic_models_all}
\end{table*}

\subsection{Expert Model Merging}
\label{subsec:model_merging}
After completing the iterative supervised refinement procedure described in \Cref{subsec:iterative_supervised_refinement} to create multiple expert models from various datasets, one path forward is to retrain a new generalist model by combining appropriate datasets based on the insights obtained from these experiments. However, we can eliminate computational redundancy by merging various expert models. This is a common practice with mature frameworks \citep{mergekit}. The literature lacks conclusive theoretical foundations for the effectiveness of model merging, but extensive experimentation has shown it is a successful strategy in practice \citep{model_merging}.

To reduce the expert merge search space, we only considered linear merges \citep{utans1996weight} of the expert models. We tested several weighting schemes based on the importance of each capability and the size of each expert's training data. In the end, our best model was obtained by assigning equal weight to each expert.

In practice, model merging reduces computational cost. However, it complicates replication and adds an additional source of potential errors.

\section{Results}
\label{sec:results}

\subsection{Arabic Language}

To measure the performance of various models in Arabic language generation and understanding, we utilized the following evaluation suite:

\begin{table*}[t]
  \centering
  \resizebox{\textwidth}{!}{%
    \begin{tabular}{lcccccc}
      \toprule
      \multirow{2}{*}{\textbf{Benchmark}} & \multirow{2}{*}{\textbf{R7B Arabic}} & \textbf{R7B} & \textbf{Gemma 9B} & \textbf{Llama 3.1 8B} & \textbf{Qwen 2.5 7B} & \textbf{Ministral 8B} \\
      & & \footnotesize\citep{cr7bblog} & \footnotesize\citep{gemma2} & \footnotesize\citep{llama3} & \footnotesize\citep{qwen2.5} & \footnotesize\citep{ministral} \\
    \midrule
      BBH \citep{bbh} & 36.2 & 36.0 & \textbf{42.1} & 29.9 & 34.9 & 25.8 \\
      MuSR \citep{musr} & \textbf{11.9} & 10.2 & 9.7 & 8.4 & 8.5 & 8.4 \\
      GPQA \citep{gpqa} & 7.9 & 7.8 & \textbf{14.8} & 2.4 & 5.5 & 4.5 \\
      MMLU Pro \citep{mmlu_pro} & 29.4 & 28.6 & 32.0 & 30.7 & \textbf{36.5} & 30.7 \\
      IfEval \citep{zhou2023instructionfollowingevaluationlargelanguage} & \textbf{83.3} & 77.1 & 74.4 & 78.6 & 75.9 & 59.0 \\
      MATH* \citep{math_leaderboard} & 19.6 & 29.9 & 19.1 & 19.3 & 50.0 & 19.6 \\
      \midrule
    Average & 31.4 & 31.6 & 32.1 & 28.2 & \textbf{35.2}& 22.0 \\
      \bottomrule 
    \multicolumn{7}{l}{\small* The MATH benchmark used in this leaderboard changed in early January due to a DMCA takedown notice for the original benchmark.} \\
    \end{tabular}%
  }
  \caption{Performance comparison of R7B Arabic against similarly sized models on multiple benchmarks. The highest score in each row is in \textbf{bold}. Command R7B Arabic retains most of the general and English capabilities of its base model, Command R7B, as indicated by the similar average scores.}
  \label{tab:evals_general}
\end{table*}

\begin{itemize}

    \item \textbf{IFEval AR}: An internal Arabic translation of the original English dataset \citep{zhou2023instructionfollowingevaluationlargelanguage} with 541 test samples. It measures a model's precise instruction following ability, with instructions such as ``use at least 300 words'' or ``do not use commas.''
    
    \item \textbf{AlGhafa-Native}: The subset\footnote{\href{https://huggingface.co/datasets/OALL/AlGhafa-Arabic-LLM-Benchmark-Native}{https://huggingface.co/datasets/OALL/AlGhafa-Arabic-LLM-Benchmark-Native}} of AlGhafa \citep{almazrouei-etal-2023-alghafa} tasks which were curated by native Arabic speakers, which encapsulates the following:
    \begin{itemize}
        \item MCQ Exams AR (562 samples) \citep{hardalov2020exams}.
        \item Belebele AR Dialects (5,400 samples) and Belebele AR MSA (900 samples) \citep{belebele}.
        \item AraFacts balanced (80 samples) \citep{arafacts}.
        \item SOQAL (155 samples) \citep{mozannar2019neuralarabicquestionanswering}.
        \item XGLUE (155 samples) \citep{xglue}.
        \item Rating sentiment no neutral (8,000 samples) and rating sentiment (6,000 samples) from the HARD-Arabic-Dataset \citep{arabic_hard_dataset}.
        \item Sentiment (1,725 samples) \citep{sentiment_dataset}.
    \end{itemize}
    We report the unweighted average percentage performance across all tasks.
    
    \item \textbf{TyDiQA-GoldP Arabic}: The 921 samples in Arabic from the original TyDiQA \cite{clark2020tydiqabenchmarkinformationseeking} golden passage (GoldP) secondary task, in which models are provided with a question and a single passage that contains the question's answer. Models are prompted to determine the substring in the passage that answers the question.
    
    \item \textbf{ArabicMMLU} \citep{koto2024arabicmmlu}: Inspired by the original MMLU \citep{mmlu} in English, ArabicMMLU is a collection of 14,575 native Arabic multiple choice questions focusing on knowledge and reasoning. It covers 40 tasks at various education levels (elementary to college) and regions (North Africa, Levant, and Gulf).
    
    \item \textbf{FaithEval Arabic}: An internal Arabic translation of a 500 sample subset of the original English dataset \cite{ming2024faithevallanguagemodelstay}. It measures the model's RAG performance when provided with unanswerable, inconsistent, or counterfactual contexts.

    \item \textbf{Multilingual ArenaHard} \citep{aya_expanse}: A machine translation of 500 questions from the original English LMArena (formerly LMSYS) Arena-Hard-Auto \citep{arena_hard} prompts into various other languages. We limit our evaluation to the Arabic subset. The evaluation uses GPT-4o as a judge to compare completions from two different models.
    
\end{itemize}

\Cref{tab:arabic_models_all} shows results compared to other models in the same size category. The Command R7B Arabic model outperforms all baselines across key Arabic benchmarks, achieving an average score of 69.3, surpassing Command R7B (65.8) and Gemma 9B (67.0). It performs at the top of its size class in the following benchmarks: Cultural Knowledge (AlGhafa-Native), Instruction Following (IFEval AR) validating our human-in-the-loop data strategy, RAG Question Answering (TyDiQA-GoldP Arabic), and RAG Faithfulness (FaithEval Arabic). In General Knowledge (ArabicMMLU), Command R7B Arabic scores third, while staying competitive with Gemma 9B and Qwen 2.7. 

\subsection{General Capabilities}

Retaining general capabilities is essential for the model to be helpful in enterprise settings. We thoroughly measured our model's performance and present the results of the standardized Hugging Face Open LLM Leaderboard benchmarks \cite{open-llm-leaderboard-v2,eval-harness}. \Cref{tab:evals_general} shows that our model excels in IfEval and MuSR, achieving the highest scores among similarly sized models. Notably, it outperforms the initial checkpoint on all benchmarks except for MATH, possibly due to the change in methodology.

These benchmark results (Table \ref{tab:arabic_models_all} and Table \ref{tab:evals_general}), coupled with auto win-rate data (\Cref{fig:model_head_to_head}), validate that our approach effectively enhances Arabic language capabilities while maintaining robust performance in enterprise applications.

\section{Conclusion}
In this work, we rapidly iterated to develop Command R7B Arabic, a small, yet competent Arabic LLM optimized for enterprise applications. By leveraging synthetic data generation, multilingual arbitrage, and human-in-the-loop interventions, we significantly improved instruction following, retrieval-augmented generation (RAG), and question answering capabilities in Arabic. However, transferring knowledge from English-centric datasets to Arabic remains an open challenge. Future work should explore more effective adaptation strategies, ensuring higher linguistic and factual alignment across languages.

\section{Limitations}
Our work focuses on Modern Standard Arabic (MSA), which is widely used in formal and professional settings but differs significantly from spoken dialects across the Arabic-speaking world. While MSA provides a strong foundation for enterprise applications, real world use cases often involve dialectal Arabic, which varies by region and context. Future work should explore dialect adaptation strategies to improve robustness across diverse Arabic varieties.

We adapted Faithfulness (FaithEval Arabic), Question Answering (TyDi QA Arabic), and Instruction Following (IFEval AR) to measure enterprise-relevant capabilities. Still, these benchmarks remain proxies rather than direct tests of real-world deployment challenges. The effectiveness of our model in enterprise workflows can only be fully validated through real-world deployment and user feedback.

\section{Acknowledgments}
This work was a collaboration between many teams in Cohere. We would like to particularly acknowledge the following people who supported the project through advice and maintenance of our core infrastructure:

Modeling Team:
Théo Dehaze,
Jesse Willman,
Lewis Stott,
Florian Strub,
Jay Alammar,
Matthias Gallé,
Samuel Cahyawijaya,
Alexandre Bérard,
Wei-Yin Ko,
Kocmi Tom,
Dennis Aumiller,
Nathan Grinsztajn,
Phil Blunsom,
Jon Ander Campos,
Yi Chern Tan,
Sander Land,
Nithya Govindarajan,
Nick Jakobi,
Adrien Morisot,
Olivia Markham;

C4AI:
Sungjin Hong,
Alejandro Salamanca,
Marzieh Fadaee,
Ahmet Üstün,
Sara Hooker;

Infrastructure:
Cécile Robert-Michon,
Jessica Xie,
Adi Bongale,
Ace Eldeib,
Sudip Roy,
Manoj Govindassamy,
Maxime Brunet,
Jeremy Pekmez,
Terrence Zhao,
Renjie Huang;

Applied ML Team:
Neeral Beladia,
Gokce Keskin,
Utsav Garg,
Jason Jung,
Hemangani Nagarajan,
Sanal Shivaprasad,
Sam Passaglia,
Edmond Wen,
Trushant Kalyanpur,
Vivek Muppalla,
Evren Tumer,
Harri Bell-Thomas;

Annotators:
Arwa Alaya,
Noha Shehata ,
Eyas Shanaah ,
Abdullah Omran,
Nermeen Isaac,
Izzat Homsi,
Mahmoud Mansour,
Mayar Soliman,
Israr Wahid,
Vanessa Choueiry,
Mona Knobloch,
Fatima Zahra Zyad;

Annotator Operations:
Claire Cheng,
Trisha Starostina,
Brenda Malacara Lopez;

Leadership:
Aidan Gomez,
Martin Kon,
Saurabh Baji,
Phil Blunsom;

External partners:
Neha Sengupta,
Ali El Filali.

\newpage
\bibliography{custom}

\clearpage  
\appendix

\end{document}